\documentclass{article}

\usepackage{microtype}
\usepackage{graphicx}
\usepackage{subfigure}
\usepackage{booktabs} 

\usepackage[T1]{fontenc}
\usepackage{lmodern}
\usepackage{hyperref}
\usepackage{url}
\usepackage{graphicx}
\usepackage{multirow}
\usepackage{amsthm}

\usepackage{amsmath}

\usepackage[utf8]{inputenc} 
\usepackage[T1]{fontenc}    
\usepackage{hyperref}       
\usepackage{url}            
\usepackage{booktabs}       
\usepackage{amsfonts}       
\usepackage{nicefrac}       
\usepackage{microtype}      
\usepackage{bm}
\usepackage{algorithm}
\usepackage{algorithmic}
\usepackage{amssymb}
\usepackage{graphicx}
\usepackage{subfigure}

\newcommand*{\ie}{\emph{i.e. }}
\newcommand*{\eg}{\emph{e.g. }}
\newcommand*{\tran}{{^{\mkern-1.5mu\mathsf{T}}}}
\newcommand*{\Diag}{{\mathrm{Diag}}}
\newcommand*{\U}{{\bm{U}}}
\renewcommand{\algorithmicrequire}{\textbf{Input:}}  
\renewcommand{\algorithmicensure}{\textbf{Output:}}

\usepackage{hyperref}

\usepackage[accepted]{icml2021}

\icmltitlerunning{Differentiable Dynamic Quantization with Mixed Precision and Adaptive Resolution}
\newcommand{\etal}{\textit{et al.}}

\begin{document}

\twocolumn[
\icmltitle{Differentiable Dynamic Quantization with Mixed Precision and\\ Adaptive Resolution}




\begin{icmlauthorlist}
\icmlauthor{Zhaoyang Zhang}{cuhk}
\icmlauthor{ Wenqi Shao}{cuhk}
\icmlauthor{ Jinwei Gu}{sb,sh}
\icmlauthor{ Xiaogang Wang}{cuhk}
\icmlauthor{ Ping Luo}{hku}
\end{icmlauthorlist}

\icmlaffiliation{cuhk}{The Chinese University of Hong Kong}
\icmlaffiliation{hku}{The University of Hong Kong }
\icmlaffiliation{sb}{SenseBrain, Ltd}
\icmlaffiliation{sh}{Shanghai AI Lab}

\icmlcorrespondingauthor{Zhaoyang Zhang}{zhaoyangzhang@link.cuhk.edu.hk}


\vskip 0.3in
]



\printAffiliationsAndNotice{}  

\begin{abstract}
Model quantization is challenging due to many tedious hyper-parameters such as precision (bitwidth), dynamic range (minimum and maximum discrete values) and stepsize (interval between discrete values). 
Unlike prior arts that carefully tune these values, we present a fully differentiable approach to learn all of them, named Differentiable Dynamic Quantization (DDQ), which has several  benefits. 
(1) DDQ is able to quantize challenging lightweight architectures like MobileNets, where different layers prefer different quantization parameters.
%
(2) DDQ is hardware-friendly and can be easily implemented using low-precision matrix-vector multiplication, making it  capable in many hardware such as ARM.
(3) 
DDQ reduces training runtime by 25\% compared to state-of-the-arts.
Extensive experiments show that DDQ outperforms prior arts on many networks and benchmarks, especially when models are already efficient and compact. \eg DDQ is the first approach that achieves lossless 4-bit quantization for MobileNetV2 on ImageNet.

\end{abstract}

\begin{figure*}
\centering
      \includegraphics[width=0.8\linewidth]{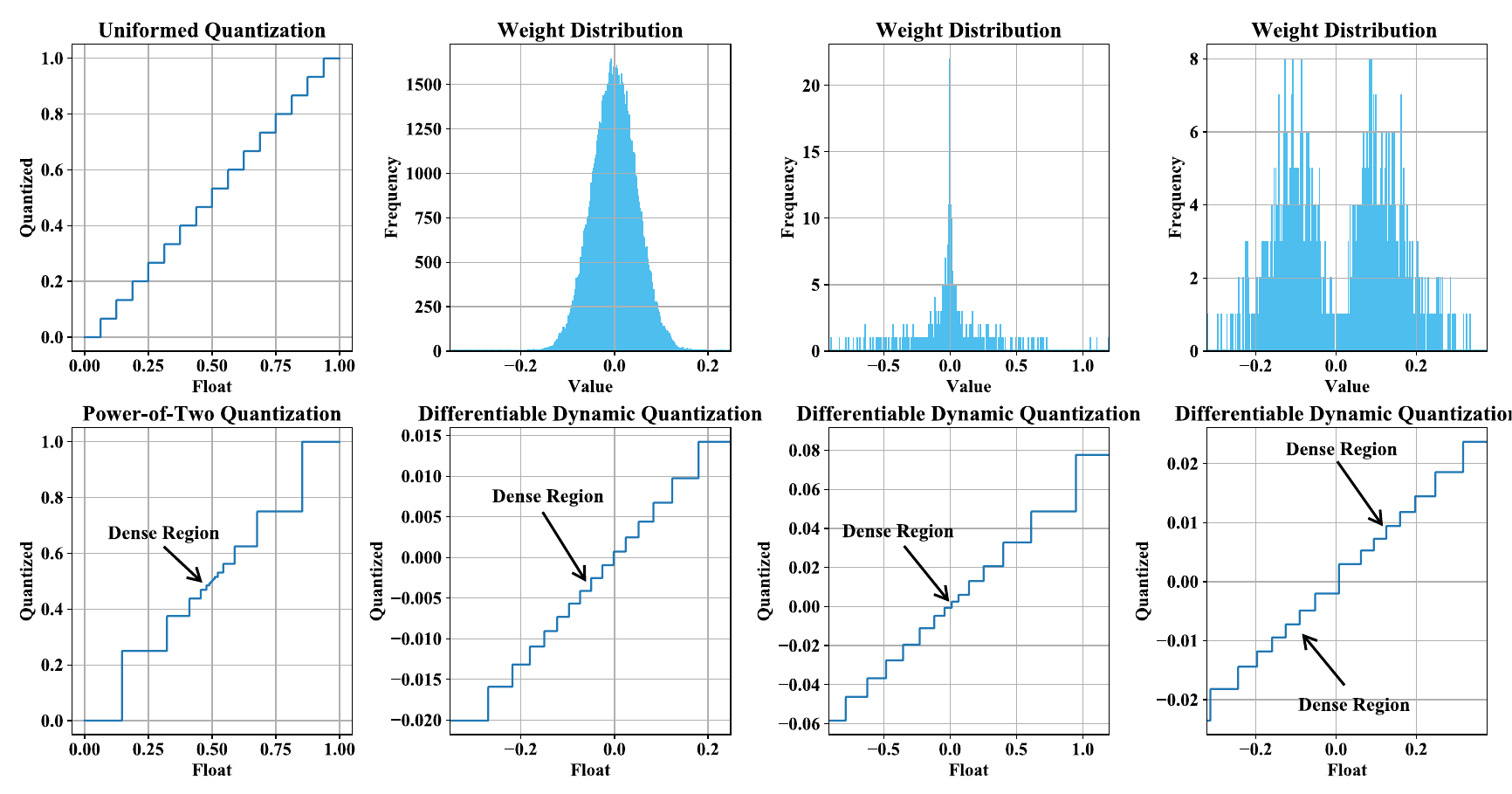} \\
   { \hspace{0.05\linewidth} (a)  \hspace{0.2\linewidth} (b) \hspace{0.18\linewidth} (c) \hspace{0.18\linewidth} (d) }
  \vspace{-7pt}
  \caption{ {\textbf{Comparisons} between methods in 4-bit quantization. \textbf{(a)} compares quantization levels between uniform and nonuniform (power-of-two) \citep{miyashita2016convolutional, zhou2017incremental,liss2018efficient,zhang2018lq} quantizers, where $x$- and $y$-axis  represent  values before and after quantization respectively\footnotemark[4] (the float values are scaled between 0 and 1 for illustration). 
  We highlight the dense region with higher ``resolution'' by arrows. We see that there is no dense region in uniform quantization (top), because the intervals between levels are the same, while a single dense region in power-of-two quantization (bottom). \textbf{(b)}, \textbf{(c)} and \textbf{ (d) } show the distributions of network weights of different layers in a MobileNetv2 \citep{sandler2018mobilenetv2} trained on ImageNet \citep{russakovsky2015imagenet}, and the corresponding quantization levels learned by our DDQ.
  We see that different layers of MobileNetv2 prefer different parameters, \eg weight distributions are Gaussian-like in (b), heavy-tailed in (c), and two-peak bell-shaped in (d). DDQ learns different quantization levels with different number of dense regions to capture these distributions. 
  }}
  \label{fig:1}
\vspace{-13pt}
\end{figure*}

\section{Introduction} 
Deep Neural Networks (DNNs) have made significant progress in many applications. However,  large memory and computations impede deployment of DNNs on portable devices.
Model quantization \citep{courbariaux2015binaryconnect, courbariaux2016binarized, zhu2016trained} that discretizes the weights and activations of a DNN becomes an important topic, but it is challenging because of two aspects. Firstly, different network architectures allocate different memory and computational complexity in different layers, making quantization suboptimal when the quantization parameters such as bitwidth, dynamic range and stepsize are freezed in each layer.
Secondly, gradient-based training of low-bitwidth models is difficult \citep{bengio2013estimating}, because the gradient of previous quantization function may vanish, \ie back-propagation through a quantized DNN may return zero gradients.

%
%


Previous approaches typically use the round operation with predefined quantization parameters, which can be summarized below. Let $x$ and $x_q$ be values before and after quantization, we have
$x_q =\mathrm{sign}(x)\cdot d\cdot \mathcal{F} (\lfloor |x|/d+0.5 \rceil)$,
where $|\cdot|$ denotes the absolute value, $\mathrm{sign}(\cdot)$ returns the sign of $x$, $d$ is the stepsize (\ie the interval between two adjacent discrete values after quantization), and $\lfloor\cdot\rceil$ denotes the round function\footnote[2]{The round function returns the closest discrete value given a continuous value.}. 
Moreover, $\mathcal{F}(\cdot)$ is a function that maps a rounded value to a desired quantization level (\ie a desired discrete value). For instance, the above equation would represent a uniform quantizer\footnote[3]{A uniform quantizer has uniform quantization levels, which means the stepsizes between any two adjacent discrete values are the same.}  when $\mathcal{F}$ is an identity mapping, or it would represent a power-of-two (nonuniform) quantizer \citep{miyashita2016convolutional, zhou2017incremental,liss2018efficient, zhang2018lq, li2019additive} when $\mathcal{F}$ is a function of power of two.

Although using the round function is simple, model accuracy would drop  when applying it on lightweight architectures such as MobileNets \citep{howard2019searching,sandler2018mobilenetv2} as observed by \cite{jain2019trained, gong2019differentiable}, because such models that are already efficient and compact do not have much room for quantization. 
%
%
As shown in Fig.\ref{fig:1}, small model is highly optimized for both efficiency and accuracy, making different layers prefer different quantization parameters.
Applying predefined quantizers to them will result in large quantization errors (\ie $\| x - x_q \|_2$),  which significantly decreases  model accuracy compared to their full-precision counterparts even after finetuning the network weights. 

%
To improve model accuracy, recent quantizers reduce $||x - x_q||_2$. 
For example, TensorRT~\citep{migacz20178}, FAQ~\citep{mckinstry2018discovering}, PACT~\citep{choi2018pact:} and TQT~\citep{jain2019trained} optimize an additional parameter that represents the dynamic range to calibrate the quantization levels, in order to better fit the distributions of the full-precision network.
%
Besides, prior arts  \citep{miyashita2016convolutional, zhou2017incremental,liss2018efficient, zhang2018lq, li2019additive} also adopted {non-uniform} quantization levels with different step-sizes.

 \footnotetext[4]{For example, a continuous value is mapped to its nearest discrete value. }

Despite the above works may reduce gap  between values before and after quantization, they 
have different 
assumptions that may not be applied to efficient networks.
For example, one key assumption is that the network weights would follow a bell-shaped distribution. However, this is not always plausible in many architectures such as MobileNets~\citep{sandler2018mobilenetv2, howard2017mobilenets, howard2019searching}, EfficientNets~\citep{tan2019efficientnet}, and  ShuffleNets~\citep{zhang2018shufflenet, ma2018shufflenet}.
For instance, Fig.\ref{fig:1}(b-d) plot different weight distributions of a MobileNetv2~\citep{sandler2018mobilenetv2} trained on ImageNet, which is a representative efficient model on embedded devices.
We find that they have irregular forms, especially when depthwise or group convolution~\citep{xie2017aggregated, huang2018condensenet, zhang2017interleaved, nagel2019data, park2020profit} are used to improve computational efficiency.
Although this problem has been identified in~\citep{krishnamoorthi2018quantizing,jain2019trained, goncharenko2019fast}, showing that per-channel scaling or bias correction could compensate some of the problems,
however, none of them could bridge the accuracy gap between the low-precision (\eg 4-bit) and full-precision models. 
We provide full comparisons with previous works in Appendix B.

To address the above issues, we contribute by  proposing Differentiable Dynamic Quantization (DDQ), to differentiablely learn quantization parameters for different network layers, including  bitwidth, quantization level, and dynamic range. 
DDQ has appealing benefits that prior arts may not have. 
%
(1) DDQ is applicable in efficient and compact networks, where different layers are quantized by different parameters in order to achieve high accuracy.
%
(2) DDQ is hardware-friendly and can be easily implemented using a low-precision matrix-vector multiplication (GEMM), making it capable in  hardware such as ARM. Moreover, a matrix re-parameterization method  is devised to reduce the matrix size  from $O(2^{2b})$ to $O(\log2^b)$, where $b$ is the number of bits.
(3) Extensive experiments show that DDQ outperforms prior arts on many networks on ImageNet and CIFAR10, especially for the challenging small models \eg MobileNets. 
To our knowledge, it is the first time to achieve lossless accuracy when quantizing MobileNet with 4 bits.  For instance, MobileNetv2+DDQ achieves comparable top-1 accuracy on ImageNet, \ie $71.9\%$ {v.s.} 71.9\% (full-precision), while ResNet18+DDQ improves the full-precision model, \ie 71.3\% {v.s.} 70.5\%.

\section{Our Approach}


\subsection{Preliminary and Notations}
In network quantization, each continuous value $x\in\mathbb{R}$ is discretized to $x_q$, which is an element from a set of discrete values. This set is denoted as a vector $\bm{q}=[q_1,q_2,\cdots,q_n]\tran$ (termed quantization levels). We have $x_q\in\bm{q}$ and $n=2^b$ where $b$ is the bitwidth. 
Existing methods often represent $\bm{q}$ using uniform or powers-of-two distribution introduced below.

\textbf{Uniform Quantization.} The quantization levels $\bm{q}_u$ of a symmetric $b$-bit uniform quantizer \citep{uhlich2020mixed} is
\begin{equation}\label{eq:uniform-quant}
\resizebox{.9\hsize}{!}{ $%
 \bm{q}_u(\bm{\theta}) = \left[-1,\cdots, \frac{-1}{2^{b-1}-1}, -0,+0,\frac{1}{2^{b-1}-1},\cdots,1\right]^{\mkern-1.0mu\mathsf{T}}\times c +\bar{x}$,}
\end{equation}

where $\bm{\theta}=\{b,c\}\tran$ denotes a set of
quantization parameters, $b$ is the bitwidth, $c$ is the clipping threshold, which represents a symmetric dynamic range\footnote[6]{In Eqn.(\ref{eq:uniform-quant}), the dynamic range is $[-c,c]$.}, and $\bar{x}$ is a constant scalar (a bias) used to shift $\bm{q}_u$\footnote[7]{Note that `$0$' appears twice in order to assure that $\bm{q}_u$ is of size $2^b$.}.
For example, $\bm{q}_u(\bm{\theta})$ is shown in the upper plot of Fig.1(a) when $c=0.5$ and $\bar{x}=0.5$. Although uniform quantization could be simple and effective,
it assumes the weight distribution is uniform that is implausible in many recent DNNs.


\textbf{Powers-of-Two Quantization.}  The quantization levels $\bm{q}_p$ of a symmetric $b$-bit powers-of-two quantizer \citep{miyashita2016convolutional, liss2018efficient} is
\begin{equation}\label{eq:power-quant}
\resizebox{.9\hsize}{!}{$%
\bm{q}_p(\bm{\theta}) = \left[-2^{-1},\cdots,-2^{-2^{b-1}+1},-0,+0,2^{-2^{b-1}+1},\cdots,2^{-1}\right]^{\mkern-1.0mu\mathsf{T}}\times c +\bar{x}.$}
\end{equation}
As shown in the bottom plot of Fig.1(a) when $c=1$ and $\bar{x}=0.5$, $\bm{q}_p(\bm{\theta})$ has a single dense region that may capture a single-peak bell-shaped weight distribution.  Power-of-two quantization can also capture  multiple-peak distribution by using the predefined additive scheme \citep{li2019additive}.


In Eqn.(\ref{eq:uniform-quant}-\ref{eq:power-quant}), both uniform and power-of-two quantizers would fix $\bm{q}$ and optimize $\bm{\theta}$, which contains the clipping threshold $c$ and the stepsize denoted as $d=1/(2^{b}-1)$~\citep{uhlich2020mixed}.
Although they learn the dynamic range and stepsize,
they have an obvious drawback, that is, the predefined quantization levels cannot fit varied distributions of weights or activations for each layer during training.


\subsection{Dynamic Differentiable Quantization (DDQ)}
\label{sec:DDQ}

\begin{figure*}
\centering
\includegraphics[width=.95\linewidth]{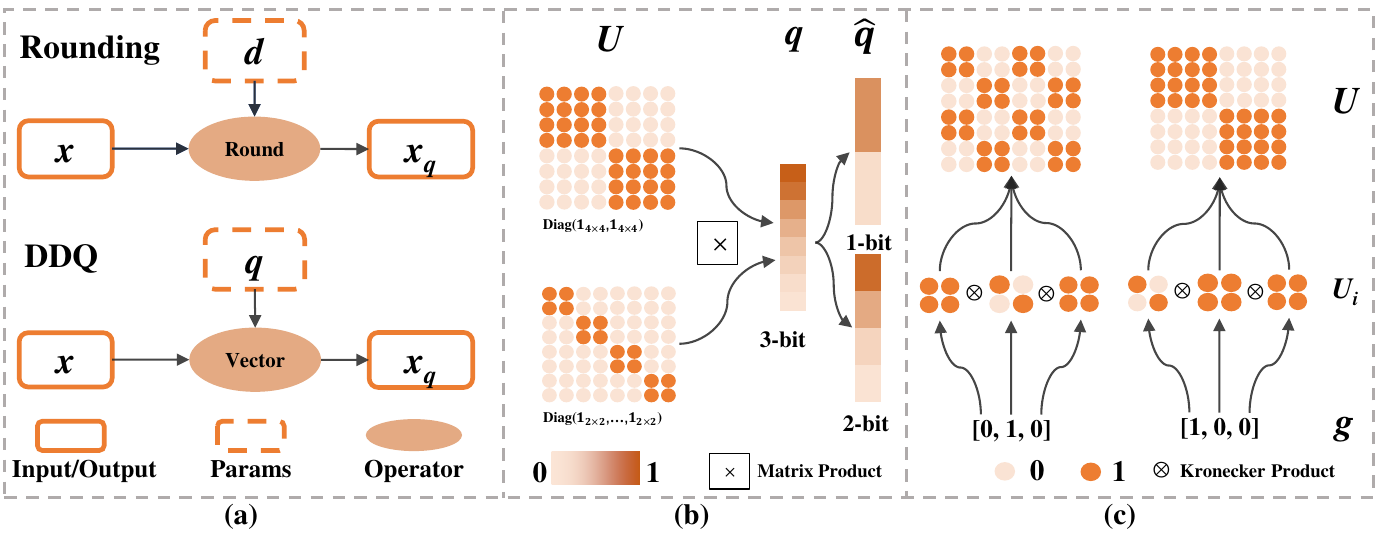}
\caption{ {\textbf{Illustrations of DDQ.} \textbf{(a)} compares computations of DDQ with the round operator. Unlike rounding methods that only learn the stepsize $d$,
DDQ treats $\bm{q}$ as trainable variable, learning arbitrary quantization levels.
\textbf{(b)} illustrates that DDQ enables mixed-precision training by using different binary block-diagonal matrix $\U$. The circles in light and dark indicate ‘0’ and ‘1’ respectively. For example, when $\bm{q}$ is of length 8 entries (\ie 3-bit) and $\U=\Diag\left(\bm{1}_{2\times 2},\cdots,\bm{1}_{2\times 2}\right)$, where $\Diag (\cdot)$ returns a matrix with the desired diagonal blocks while the off-diagonal blocks are zeros and $\bm{1}_{2\times 2}$ denotes a 2-by-2 matrix of ones, we have $\hat{\bm{q}}=\U\tran{\bm{q}}$ that enables DDQ to represent a 2-bit quantizer by averaging neighboring discrete values in $\bm{q}$.
Another example is a 1-bit quantizer when $\U=\Diag\left(\bm{1}_{4\times 4},\bm{1}_{4\times 4}\right)$. \textbf{(c)} shows relationship between gating variables $\bm{g}=\{g_i\}_{i=1}^b$ and $\U$. For example, when the entries of $\bm{g}=[1,0,0]$ are arranged in an descending order and let $s=\sum_{i=1}^bg_i=1+0+0=1$, $\U$ has $2^s=2^1=2$ number of all-one diagonal blocks. In such case, DDQ is a $s=1$ bit quantizer.
}}\label{fig:2}
\end{figure*}

\textbf{Formulation for Arbitrary Quantizaion.} Instead of freezing the quantization levels, DDQ learns all quantization hyperparameters. 
Let $Q(x;\bm{\theta})$ be a function with a set of parameters $\bm{\theta}$ and $x_q=Q(x;\bm{\theta})$ turns a continuous value $x$ into an element of $\bm{q}$ denoted as $x_q\in\bm{q}$, where $\bm{q}$ is initialized as a uniform quantizer and can be implemented in low-precision values according to hardware's requirement.
DDQ is formulated by low-precision matrix-vector product,
\begin{equation}\label{eq:DQ}
\resizebox{.95\hsize}{!}{ $%
    x_q= {\bm{q}}\tran \frac{\U}{{Z}_U}\bm{x}_o,\,\mathrm{where}\,\,{x}^i_o=\left\{
\begin{array}{l}
1\quad \mathrm{if}\, i=\mathrm{argmin}_j|\frac{1}{Z_U}(\bm{U}\tran\bm{q})_j-x|\\
0\quad \mathrm{otherwise}
\end{array},
\right.$ }
\end{equation}
where ${x}^i_o\in\bm{x}_o$, $\bm{x}_o\in\{0,1\}^{n\times 1}$ denotes  a binary vector that has only one entry of `1' while others are `0', in order to select one quantization level from ${\bm{q}}$ for the continuous value $x$. 
Note that we reparameterize $\bm{q}$ by a trainable vector $\tilde{\bm{q}}$, such that 
\begin{equation}
\resizebox{.8\hsize}{!}{$%
    \bm{q}=R(\tilde{\bm{q}})(x_{max}-x_{min})/(2^b-1)+x_{min}, $ }
\end{equation}
where $R()$ denotes a uniform quantization function transforming $\tilde{q}$ to given $b_q$ bits ($b_q < b$).
%
Eqn.(\ref{eq:DQ}) has parameters $\bm{\theta}=\{\tilde{\bm{q}} ,\U\}$, which are trainable by stochastic gradient descent (SGD), 
making DDQ automatically capture weight distributions of the full-precision models as shown in Fig.\ref{fig:1}(b-d). 
Here $\U\in\{0,1\}^{n\times n}$ is a binary block-diagonal matrix and $Z_U$ is a constant normalizing factor used to average the discrete values in $\bm{q}$ in order to learn bitwidth.  Intuitively, different values of $\bm{x}_o$, $\U$ and $\bm{q}$ make DDQ represent different quantization approaches as discussed below. 
To ease understanding, Fig.\ref{fig:2}(a) compares the computational graph of DDQ with the rounding-based methods. We see that DDQ learns the entire quantization levels instead of just the stepsize $d$ as prior arts did.


\textbf{Discussions of Representation Capacity.} DDQ represents a wide range of quantization methods. For example, when $\bm{q}=\bm{q}_u$ (Eqn.(\ref{eq:uniform-quant})), $Z_U=1$, and $\U=\bm{I}$ where $\bm{I}$ is an identity matrix, Eqn.(\ref{eq:DQ}) represents an ordinary uniform quantizer. When $\bm{q}=\bm{q}_p$ (Eqn.(\ref{eq:power-quant})), $Z_U=1$, and $\U=\bm{I}$, Eqn.(\ref{eq:DQ}) becomes a power-of-two quantizer. When $\bm{q}$ is learned, it represents arbitrary quantization levels with different dense regions.

Moreover, DDQ enables mixed precision training when $\U$ is block-diagonal. For example, as shown in Fig.\ref{fig:2}(b), when $\bm{q}$ has length of 8  entries (\ie 3-bit), $Z_U=\frac{1}{2}$, and $\U=\Diag\left(\bm{1}_{2\times 2},\cdots,\bm{1}_{2\times 2}\right)$, where $\Diag (\cdot)$ returns a matrix with the desired diagonal blocks and its off-diagonal blocks are zeros and $\bm{1}_{2\times 2}$ denotes a 2-by-2 matrix of ones, $\U$ enables Eqn.(\ref{eq:DQ}) to represent a 2-bit quantizer by averaging neighboring discrete values in $\bm{q}$.  
For another example, when $\U=\Diag\left(\bm{1}_{4\times 4},\bm{1}_{4\times 4}\right)$ and $Z_U=\frac{1}{4}$, Eqn.(\ref{eq:DQ}) turns into a 1-bit quantizer. 
Besides, when $\bm{x}_o$ is a soft one-hot vector with multiple non-zero entries,  Eqn.(\ref{eq:DQ}) represents soft quantization that one continuous value can be mapped to multiple discrete values.

\textbf{Efficient Inference on Hardware.}
\label{sec:efficient}
DDQ is a unified quantizer that supports adaptive $\bm{q}$ as well as predefined ones \eg uniform and power-of-two. It is friendly to hardware with limited resources.
As shown in Eqn.(\ref{eq:DQ}), DDQ reduces to a uniform quantizer when $\bm{q}$ is uniform.
In this case, DDQ can be efficiently computed by a rounding function as the step size is determined by $\bm{U}$ after training  (\ie don't have $\bm{U}$ and matrix-vector product when deploying in hardware like other uniform quantizers). 
In addition, DDQ with adaptive $\bm{q}$ can be implemented using low-precision general matrix multiply (GEMM).
For example, let $y$ be a neuron's activation, $y = Q(w;\bm{\theta}) x_q = {\bm{q}}\tran \frac{\bm{U}}{{Z}_U} \bm{w}_o x_q$, 
where $x_q$ is a discretized feature value, $w$ is a continuous weight parameter to be quantized, and $\bm{U}$ and $\bm{w}_o$ are binary matrix and one-hot vector of $w$ respectively. To accelerate, we can calculate the major part $\frac{\bm{U}}{Z_U} \bm{w}_o x_q$ using low-precision GEMM first and then multiplying a short 1-d vector $\bm{q}$, which is shared for all convolutional weight parameters and can be float32, float16 or INT8 given specific hardware requirement.
The latency in hardware is compared in later discussion.


\subsection{Matrix Reparameterization of \textit{U}}\label{sec:U}

In Eqn.(\ref{eq:DQ}), $\U$ is a learnable matrix variable, which is challenging to optimize in two aspects. First, to make DDQ a valid quantizer, $\U$ should have binary block-diagonal structure, which is difficult to learn by using SGD. Second, the size of $\U$ (number of parameters) increases  when the bitwidth increases \ie $2^{2b}$. 
Therefore, rather than directly optimize $\U$ in the backward propagation using SGD, we explicitly construct $\U$ by composing a sequence of small matrices in the forward propagation~\citep{luo2019differentiable}.

 

\textbf{A Kronecker Composition for Quantization.} The matrix $\U$ can be reparameterized to reduce number of parameters from $2^{2b}$ to $\log2^b$ to ease optimization. Let $\{\U_1,\U_2,\cdots,\U_b\}$ denote a set of $b$  small matrices of size 2-by-2, $\U$ can be constructed by
$\U=\U_1\otimes\U_2\otimes\cdots\otimes\U_b$,
where $\otimes$ denotes Kronecker product and each $\U_i$ ($i=1...b$) is either a 2-by-2 identity matrix (denoted as $\bm{I}$) or an all-one matrix (denoted as $\bm{1}_{2\times 2}$), making $\U$ block-diagonal after composition. 
For instance, when $b=3$, $\U_1=\U_2=\bm{I}$ and $\U_3=\bm{1}_{2\times 2}$, we have $\U=\Diag\left(\bm{1}_{2\times 2},\cdots,\bm{1}_{2\times 2}\right)$ and Eqn.(\ref{eq:DQ}) represents a 2-bit quantizer as shown in Fig.\ref{fig:2}(b). 

To pursue a more parameter-saving composition, we further parameterize each $\U_i$ by using a single trainable variable. As shown in Fig.\ref{fig:2}(c), we have
$\U_i=g_i\bm{I}+(1-g_i)\bm{1}_{2\times 2}$, where $g_i=H(\hat{g}_i)$ and $H(\cdot)$ is a Heaviside step function\footnote[8]{The Heaviside step function returns `0' for negative arguments and `1' for positive arguments.}, \ie $g_i=1$ when $\hat{g}_i\geq 0$; otherwise $g_i=0$.
Here $\{g_i\}_{i=1}^{b}$ is a set of gating variables with binary values. Intuitively, each $\U_i$ switches between a 2-by-2 identity matrix and a 2-by-2 all-one matrix.

In other words, $\U$ can
be constructed by applying a series of Kronecker products 
involving only $\bm{1}_{2\times 2}$ and $\bm{I}$. Instead of updating the entire matrix $\U$, it can be learned by only a few variables $\{\hat{g}_i\}_{i=1}^b$, significantly reducing the number of parameters from $2^b\times 2^b=2^{2b}$ to $b$. 
In summary, the parameter size to learn $\U$ is merely the number of bits. With Kronecker composition, the quantization parameters of DDQ is $\bm{\theta}=\{\bm{q},\{\hat{g}_i\}_{i=1}^b\}$, which could be different for different layers or kernels (\ie layer-wise or channel-wise quantization) and the parameter size is negligible compared to the network weights, making different layers or kernels have different quantization levels and bitwidth.


\textbf{Discussions of Relationship between $\U$ and $\bm{g}$.} Let $\bm{g}$ denote a vector of gates $[g_1,\cdots,g_b]\tran$. In general, different values of $\bm{g}$ represent different block-diagonal structures of $\U$ in two aspects. (1) \textbf{Permutation.} As shown in Fig.\ref{fig:2}(c), $\{g_i\}_{i=1}^b$ should be permuted in an descending order by using a permutation matrix. Otherwise, $\U$ is not block-diagonal when $\bm{g}$ is not ordered, making DDQ an invalid quantizer. For example, $\bm{g}=[0,1,0]$ is not ordered compared to $\bm{g}=[1,0,0]$. 
(2) \textbf{Sum of Gates.} Let $s=\sum_{i=1}^b g_i$ be the sum of gates and $0\leq s\leq b$. We see that $\U$ is a block-diagonal matrix with $2^s$ diagonal blocks, implying that $\U\tran{\bm{q}}$ has $2^s$ different discrete values and represents a $s$-bit quantizer. For instance, as shown in Fig.\ref{fig:2}(b,{c}) when $b=3$, $\bm{g}=[1,0,0]\tran$ and $\U=\Diag\left(\bm{1}_{4\times 4},\bm{1}_{4\times 4}\right)$, we have a $s=1+0+0=1$ bit quantizer. DDQ enables to regularize the value of $s$ in each layer given memory constraint, such that optimal bitwidth can be assigned to different layers of a DNN.


\section{Training DNN with DDQ}


DDQ is used to train a DNN with mixed precision to satisfy memory constraints, which reduce the memory to store the network weights and activations, making a DNN appealing for deployment in embedded devices.

\subsection{DNN with Memory Constraint} 

Consider a DNN with $L$ layers trained using DDQ, the forward propagation of each layer can be written as
\begin{equation}\label{eq:forward}
\resizebox{.85\hsize}{!}{$ \bm{y}^l={F}\big(Q(\bm{W}^l;\bm{\theta}^l)*Q(\bm{y}^{l-1})+Q(\bm{b}^l;\bm{\theta}^l)\big),\,l=1,2,\cdots,L$}
\end{equation}
where $*$ denotes convolution, $\bm{y}^{l}$ and $\bm{y}^{l-1}$ are the output and input of the $l$-th layer respectively, $F$ is a non-linear activation function such as ReLU, and $Q$ is the quantization function of DDQ. Let $\bm{W}^l\in \mathbb{R}^{C_{out}^l\times C_{in}^l\times K^l \times K^l}$ and $\bm{b}^l$ denote the  convolutional kernel and bias vector (network weights), where $C_{out}, C_{in}$, and $K$ are the output and input channel size, and the kernel size respectively.
Remind that in DDQ, $\{g_i^l\}_{i=1}^{b}$ is a set of  gates at the $l$-th layer and the bitwidth can be computed by $s^l=\sum_{i=1}^b g^l_i$. For example, the total memory footprint (denoted as $\zeta$) can be computed by
\begin{equation}\label{eq:memory}
    \zeta(s^1,\cdots,s^L)=\sum_{l=1}^L C_{out}^l C_{in}^l (K^l)^2 2^{s^l},
\end{equation}
which represents the memory to store all network weights at the $l$-th layer when the bitwidth is $s^l$.

If the desired memory is $\zeta(b^1,\cdots,b^L)$,
we could use a weighted product
to approximate the
Pareto optimal solutions to train a $b$-bit DNN. The loss function is


\begin{equation}\label{eq:objective}
\resizebox{.9\hsize}{!}{$%
\begin{aligned}
    \min_{\bm{W^l},\bm{\theta}^l}\quad \mathcal{L}(\{\bm{W}^l\}_{l=1}^L,\{\bm{\theta}^l\}_{l=1}^L)\cdot \left(\frac{\zeta(b^1,\cdots,b^L)}{\zeta(s^1,\cdots,s^L)}\right)^\alpha\quad
    \\
    \mathrm{s.t.}\,\, \zeta(s^1,\cdots,s^L)\leq\zeta(b^1,\cdots,b^L),
\end{aligned}
$}
\end{equation}

where the loss $\mathcal{L}(\cdot)$ is reweighted by the ratio between the desired memory and the practical memory similar to~\citep{tan2019mnasnet, deb2014multi}. $\alpha$ is a hyper-parameter. We have $\alpha=0$ when the memory constraint is satisfied. Otherwise, $\alpha<0$ is used to penalize the memory consumption of the network.
%

%

\begin{figure*}[htbp]
\centering
\vspace{-5pt}
\subfigure{
\includegraphics[width=0.45\linewidth]{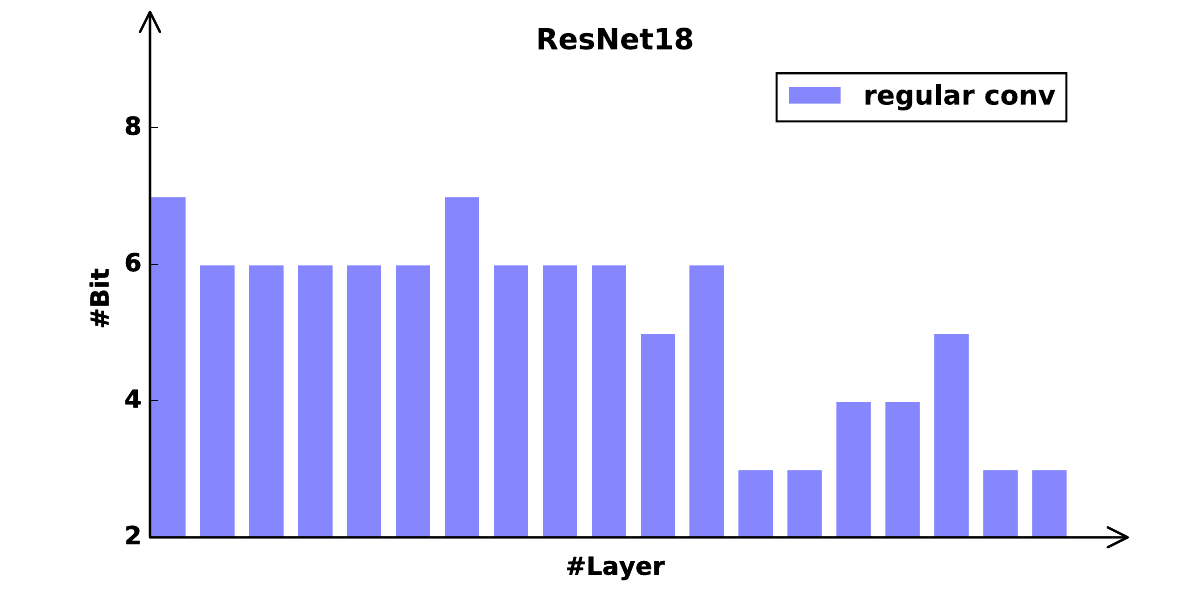}
}
\centering
\subfigure{
\includegraphics[width=0.45\linewidth]{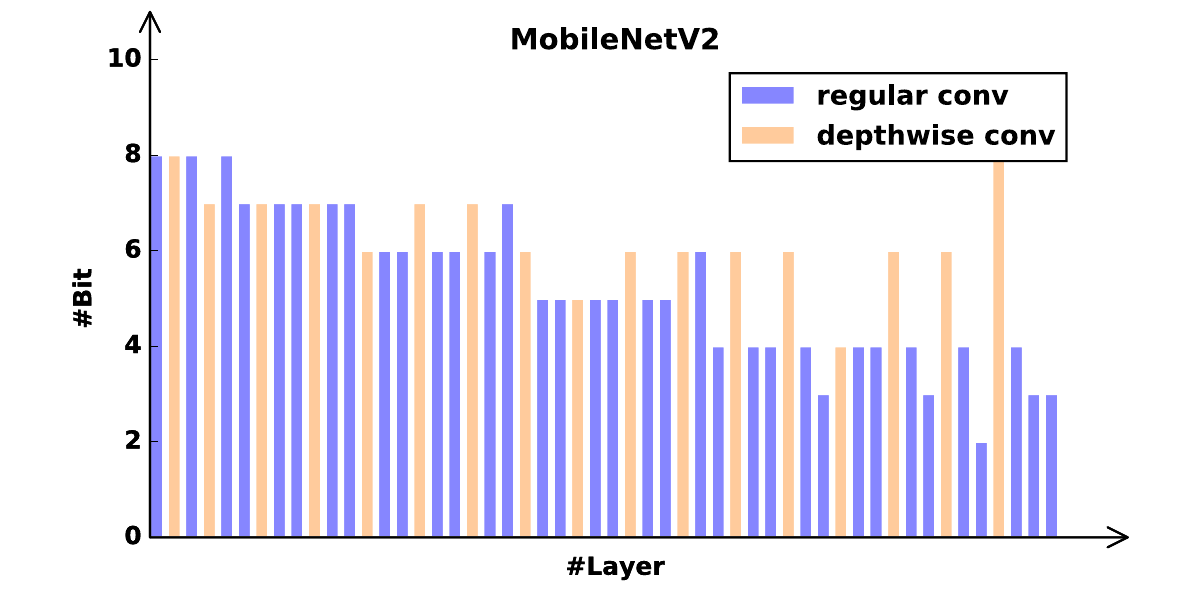}
}
\vspace{-10pt}
\caption{ {Learned quantization policy of each layer for ResNet18 and MobileNetV2 trained by DDQ on ImageNet. DDQ learns to allocate more bits to lower layers and depthwise layers of the networks.}}
\vspace{-10pt}

\label{fig:mixed}
\end{figure*}

\subsection{Updating Quantization Parameters}
\label{Sec:update}

All parameters of DDQ can be optimized by using SGD. This section derives their update rules. We omit the superscript `$l$' for simplicity.


\textbf{Gradients \emph{w.r.t.} $\bm{q}$.} To update $\bm{q}$, we reparameterize $\bm{q}$ by a trainable vector $\tilde{\bm{q}}$, such that 
$\bm{q}=R(\tilde{\bm{q}})(x_{max}-x_{min})/(2^b-1)+x_{min}$, in order to make each quantization level lies in $[x_{min},x_{max}]$, where $x_{max}$ and $x_{min}$ are the maximum and minimum continuous values of a layer, and $R()$ denotes a uniform quantization function transforming $\tilde{q}$ to given $b_q$ bits ($b_q < b$). 
%
Let $\bm{x}$ and $\bm{x}_q$ be two vectors stacking values before and after quantization respectively (\ie $x$ and $x_q$),
the gradient of the loss function with respect to each entry $q_k$ of $\bm{q}$ is given by
\begin{equation}\label{eq:grad-q}
\frac{\partial \mathcal{L}}{\partial q_k}=\sum_{i=1}^{N}\frac{\partial \mathcal{L}}{\partial {x}_q^{i}}\frac{\partial {x}_q^{i}}{\partial q_k}=\frac{1}{Z_U}\sum_{i\in \mathcal{S}_k}\frac{\partial \mathcal{L}}{\partial {x}_q^{i}},
\vspace{-2pt}
\end{equation}
where  ${x}_q^{i}=Q({x}^{i};\bm{\theta})$ is the output of DDQ quantizer and
$\mathcal{S}_k$ represents a set of indexes of the values discretized to the corresponding quantization level $q_k$. 
In Eqn.(\ref{eq:grad-q}), we see that the gradient with respect to the quantization level $q_k$ is the summation of the gradients $\frac{\partial \mathcal{L}}{\partial {x}_q^{i}}$.
In other words, the quantization level  in denser regions would have larger gradients. The gradient \textit{w.r.t.} gate variables $\{g_i\}_{i=1}^b$ are discussed in Appendix A.



\textbf{Gradient Correction for $\bm{x}_q$.} In order to reduce the quantization error $\|\bm{x}_q-\bm{x}\|_2^2$, a gradient correction term is proposed to regularize the gradient with respect to the quantized values,
\begin{equation}\label{eq:grad-corr}
\frac{\partial \mathcal{L}}{\partial \bm{x}}\leftarrow \frac{\partial \mathcal{L}}{\partial \bm{x}_q},\quad \frac{\partial \mathcal{L}}{\partial \bm{x}_q}\leftarrow \frac{\partial \mathcal{L}}{\partial \bm{x}_q}+ \lambda(\bm{x}_q-\bm{x}),
\end{equation}
where the first equation holds by applying STE. In Eqn.(\ref{eq:grad-corr}), we first assign the gradient of $\bm{x}_q$ to that of $\bm{x}$ and then add a  correction term $\lambda(\bm{x}_q-\bm{x})$. In this way, the corrected gradient can be back-propagated to the quantization parameters ${\bm{q}}$ and $\{g_i\}_{i=1}^b$ in Eqn.(\ref{eq:grad-q}), while not affecting the gradient of $\bm{x}$.
Intuitively,
this gradient correction term is effective and can be deemed as the $\ell_2$ penalty on $\|\bm{x}-\bm{x}_q\|_2^2$. 
Please note that this is not equivalent to simply impose a $\ell_2$ regularization directly on the loss function, which would have  no effect when STE is presented.

\begin{algorithm}[htb] 
\caption{ Training procedure of DDQ} 
\label{alg:one} 
\begin{algorithmic}[1] 
\REQUIRE  the full precision kernel $\bm{W}$ and bias kernel $\bm{b}$, quantization parameter $\bm{\theta}=\{\bm{q},\{\hat{g}_i\}_{i=1}^b\}$, the target bitwidth of $b_m$, input activation $\bm{y}_{in}$.\\
\ENSURE the output activation $\bm{y}_{out}$\\
\STATE Apply DDQ to the kernel $\bm{W}$, bias $\bm{b}$, input activation $\bm{y}_{in}$ by Eqn.(3)
\STATE Compute the output activation $\bm{y}_{out}$ by Eqn.(4) 
\label{code:fram:add}
\STATE Compute the loss $\mathcal{L}$ by Eqn.(6) and gradients $\frac{\partial \mathcal{L}}{\partial \bm{y}_{out}}$ 
\label{code:fram:classify}
\STATE Compute the gradient of ordinary kernel weights and bias $\frac{\partial \mathcal{L}}{\partial {\bm{W}}}$, $\frac{\partial \mathcal{L}}{\partial {\bm{m}}}$
\STATE Applying gradient correction in Eqn.(9) to compute the gradient of parameters, $\frac{\partial \mathcal{L}}{\partial \bm{q}}$,  $\frac{\partial \mathcal{L}}{\partial \hat{g}_{i}}$ by Eqn.(7) and Eqn.(8) 
\STATE Update $\bm{W},\bm{m}, \hat{g}_{i}$ and $\bm{q}$.
\end{algorithmic}
\end{algorithm}

\textbf{Implementation Details.}
Training DDQ can be simply implemented in existing  platforms such as PyTorch and Tensorflow. The forward propagation only involves differentiable functions except 
the Heaviside step function. In practice, STE can be utilized to compute its gradients, \ie $\frac{\partial x_q}{\partial x} =1_{\hat{q}_{min}\leq x\leq \hat{q}_{max}}$  and $\frac{\partial g_k}{\partial \hat{g}_k}=1_{|\hat{g}_k|\leq 1}$, where $\hat{q}_{min}$ and $\hat{q}_{max}$ are minimum and maximum value in $\hat{\bm{q}}=\U\tran\bm{q}$. Algorithm.~\ref{alg:one} provides detailed procedure. The codes will be released.

\vspace{-8pt}
\section{Experiments}

We extensively compare DDQ with existing state-of-the-art methods and conduct multiple ablation studies on ImageNet~\citep{russakovsky2015imagenet} and CIFAR dataset~\citep{krizhevsky2009learning} (See Appendix D for evaluation on CIFAR). 
The reported validation accuracy is simulated with $b_q=8$, if no other states. 

\begin{table*}[h]
\centering

\scalebox{0.7}
{
        \begin{tabular}{l |c |c c c  |c c c} \hline 
        \toprule
        \multicolumn{2}{c}{} & \multicolumn{3}{c}{ MobileNetV2}    &  \multicolumn{3}{c}{ResNet18}              \\
         \cmidrule(r){3-5}         \cmidrule(r){6-8} 
        & Training Epochs & Model Size & Bitwidth (W/A) & Top-1 Accuracy & Model Size & Bitwidth (W/A) & Top-1 Accuracy \\
        \hline
        Full Precision & 120 &  13.2 MB & 32 & 71.9 &  44.6 MB & 32 & 70.5 \\ \hline
        
        \textbf{DDQ (ours)} & 30&  1.8 MB & 4 / 8 (mixed) &\textbf{ 71.7} & 5.8 MB & 4 / 8 (mixed) & \textbf{71.0} \\ 
        \textbf{DDQ (ours)} & 90 &  1.8 MB & 4 / 8 (mixed) &\textbf{ 71.9} & 5.8 MB & 4 / 8 (mixed) & \textbf{71.3} \\
                
        Deep Compression~\citep{han2015deep}& - & 1.8 MB & 4 / 32  & 71.2 & - & - & - \\
        HMQ~\citep{habi2020hmq}   & 50 & 1.7 MB & 4 / 32 (mixed) & 71.4 & - & - & - \\ 
        HAQ~\citep{wang2019haq} & 100 & 1.8 MB & 4  / 32 (mixed)  & 71.4 & - & - & - \\
        WPRN~\citep{mishra2017wrpn}& 100 & - & - & - & 5.8 MB & 4 / 8 & 66.4  \\
        BCGD~\citep{baskin2018nice}& 80 & - & - & - & 5.8 MB & 4 / 8 & 68.9  \\
        LQ-Net~\citep{zhang2018lq} & 120 & - & - & - & 5.8 MB & 4 / 32 & 70.0 \\
        \hline
        \textbf{DDQ (ours)} & 90 & 1.8 MB & 4 / 4 (mixed) & \textbf{71.8} & 5.8 MB & 4 / 4 (mixed) & \textbf{71.2} \\ 
        \textbf{DDQ (ours)} & 30 & 1.8 MB & 4 / 4 (mixed) & \textbf{71.5} & 5.8 MB & 4 / 4 (mixed) & \textbf{71.0} \\ 
        \textbf{DDQ (ours)} & 90 & 1.8 MB & 4 / 4 (fixed) & \textbf{71.3} & 5.8 MB & 4 / 4 (fixed) & \textbf{71.1} \\ 
       \textbf{DDQ (ours)} & 30 & 1.8 MB & 4 / 4 (fixed) & \textbf{70.7} & 5.8 MB & 4 / 4 (fixed) & \textbf{70.7} \\ 
        PROFIT \citep{park2020profit} & 140 & 1.8 MB & 4 / 4   & 71.5 & -  & -  & - \\ 
        SAT~\citep{jin2019towards}  & 150 & 1.8 MB  & 4 / 4 & 71.1  & - & - & -  \\
        HMQ \citep{habi2020hmq} & 50 & 1.7 MB & 4 / 4 (mixed) & 70.9 & - & - & - \\ 
        APOT \citep{li2019additive} & 30 & 1.8 MB & 4 / 4 & 69.7* & - & - & - \\ 
        APOT \citep{li2019additive} & 100 & 1.8 MB & 4 / 4   & 71.0* & 5.8 MB & 4 / 4  & 70.7 \\ 
        LSQ \citep{esser2019learned}& 90 & 1.8 MB & 4 / 4   & 70.6* & 5.8 MB & 4 / 4   & 71.1 \\

        PACT~\citep{choi2018pact:} & 110 & 1.8 MB & 4 / 4 & 61.4 & 5.6 MB & 4 / 4 & 69.2 \\
        DSQ~\citep{gong2019differentiable}& 90 & 1.8 MB & 4 / 4 & 64.8 & 5.8 MB & 4 / 4 & 69.6 \\
        TQT~\citep{jain2019trained}& 50 & 1.8 MB  & 4 / 4 & 67.8  & 5.8 MB & 4 / 4 & 69.5  \\
        Uhlich~\etal~\citep{uhlich2020mixed}& 50 & 1.6 MB  & 4 / 4 (mixed)  & 69.7  & 5.4 MB & 4 / 4  & 70.1   \\ 
        QIL~\citep{jung2019learning} & 90& & - & - & 5.8 MB & 4 / 4 & 70.1  \\
        LQ-Net~\citep{zhang2018lq} & 120& & - & - & 5.8 MB & 4 / 4 & 69.3  \\
        NICE~\citep{baskin2018nice} & 120& & - & - & 5.8 MB & 4 / 4 & 69.8  \\
        BCGD~\citep{baskin2018nice} & 80 & & - & - & 5.8 MB & 4 / 4 & 67.4  \\
        Dorefa-Net~\citep{zhou2016dorefa} & 110 &  & - & - & 5.8 MB & 4 / 4 & 68.1  \\
        \hline
        \end{tabular}
        }
        \vspace{-5pt}
        \caption{Comparisons between DDQ and state-of-the-art quantizers on ImageNet. “W/A” means bitwidth of weight and activation respectively. Mixed precision approaches are annotated as ``mixed''.  "-" denotes the absence of data in previous papers. We see that DDQ outperforms prior arts with much less training epochs.  * denotes our re-implemented results using public codes. Note that PROFIT  \citep{park2020profit} achieves 71.5\% on MobileNetv2 using a progressive training scheme (reducing bitwidth gradually from 8-bit to 5, 4-bit, 15 epochs each stage and 140 epochs in total).}
        \label{tab:comparison}
        \vspace{-15pt}
\end{table*}

\subsection{Evaluation on ImageNet}
\begin{figure}
  \begin{center}
\includegraphics[width=1\linewidth]{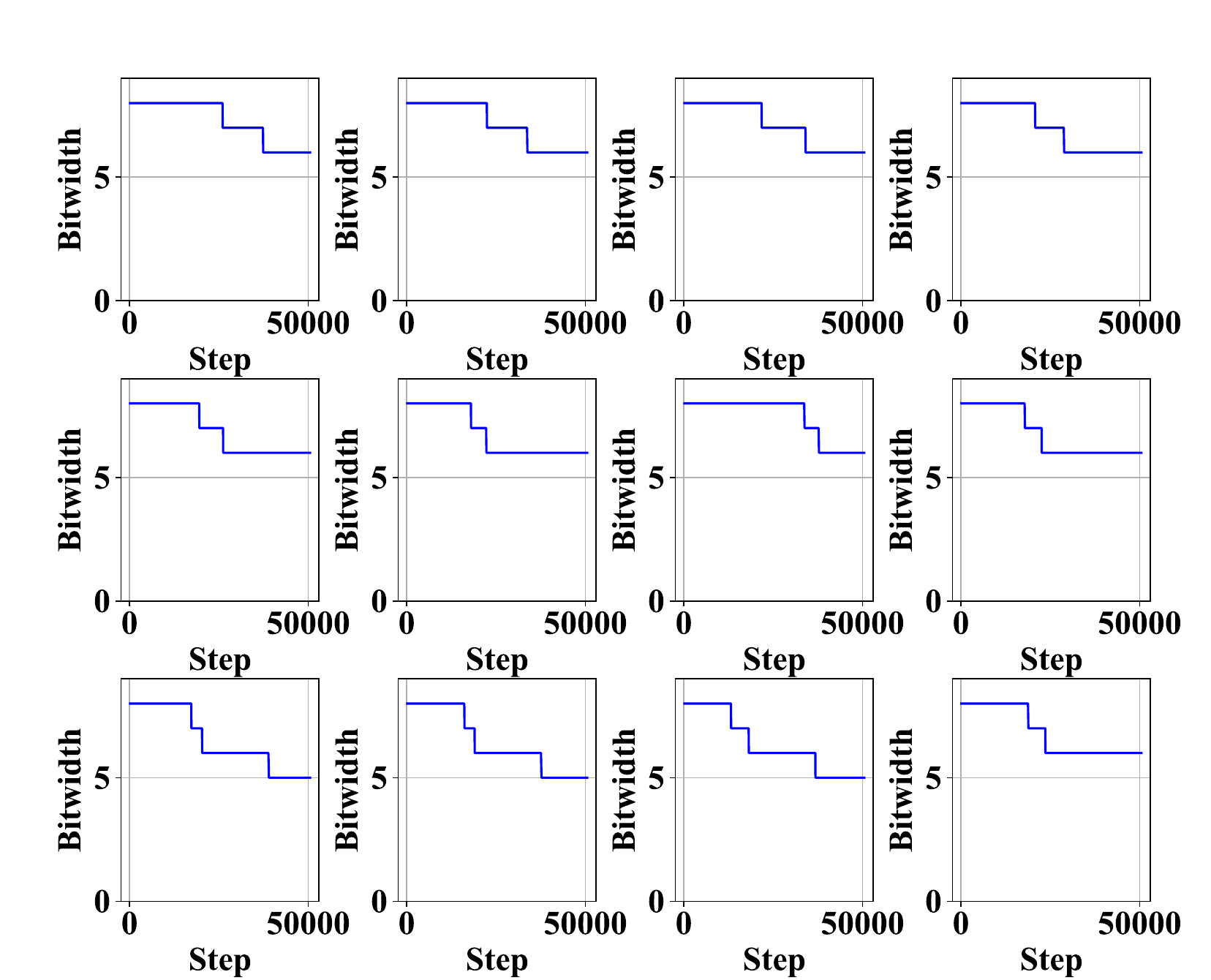}
  \end{center}
\vspace{-10pt}
\caption{  Evolution of bitwidth of top layers when training ResNet18. We can see that DDQ can learn to assign bitwidth to each layer under given memory footprint constraints. Full dynamics please see im Appendix. }
\vspace{-25pt}

\label{fig:bit}
\end{figure}

\textbf{Comparisons with Existing Methods.} Table~\ref{tab:comparison} compares DDQ with existing methods in terms of model size, bitwidth, and top1 accuracy on ImageNet using MobileNetv2 and ResNet18, which are two representative networks for portable devices. 
%
We see that DDQ outperforms recent state-of-the-art approaches by significant margins in different settings. For example, When trained for 30 epochs,  MobileNetV2+DDQ yields $71.7\%$ accuracy when quantizing weights and activations using 4 and 8 bit respectively, while achieving $71.5\%$ when training with 4/4 bit. These results
%
only drop $0.2\%$ and $0.4\%$ compared to the 32-bit full-precision model, outperforming all other quantizers, which may decrease performance a lot ($2.4\% \mathtt{\sim} 10.5\%$). 
For ResNet18, DDQ outperforms all methods even the full-precision model (\ie $71.0\%$ vs $70.5\%$). 
More importantly, DDQ is trained for $30$ epochs, reducing the training time compared to most of the reported approaches that trained much longer (\ie $90$ or $120$ epochs).  Note that PROFIT~\citep{park2020profit} achieves $71.5\%$ on MobileNetv2 using a  progressive training scheme (reducing bitwidth gradually from 8-bit to 5, 4-bit, 15 epochs each stage and 140 epochs in total.) This is quite similar to mixed-precision learning process of DDQ,  in which bitwidth of each weight is initialized to maximum bits and learn to assign proper precision to  each layer by decreasing the layerwise bitwidth. More details can see in Fig.\ref{fig:bit}. 
 


%
%

Fig.\ref{fig:mixed} shows the converged bitwidth for each  layer of MobileNetv2 and ResNet18 trained with DDQ. 
We have two interesting findings.
(1) Both networks tend to apply more bitwidth in lower layers, which have fewer parameters and thus being less regularized by the memory constraint.  This allows to learn better feature representation, alleviating the performance drop.
(2) As shown in the right hand side of Fig.\ref{fig:mixed}, 
we observe that depthwise convolution has larger bitwidth than  the regular convolution. As found in \citep{jain2019trained}, the depthwise convolution with irregular weight distributions is the main reason that makes quantizing MobileNet difficult.With mixed-precision training, DDQ allocates more bitwidth to depthwise convolution to alleviate this difficulty.
%

\begin{table}
\centering
 
\scalebox{0.7
}{
        \begin{tabular}{l |c c | c |c c }
        \hline
         & UQ & PoT  &  DDQ (fixed) & \multicolumn{2}{c}{DDQ (mixed)}   \\ \hline
         Maximum bitwidth  & 4 &  4 & 4  & 6  &  8 \\ 
         Target bitwidth & 4 & 4 & 4 & 4 & 4 \\ \hline
         Weight memory footprint & $1\times$ & $1\times$ & $1\times$ & $0.98\times$ & $1.03\times$ \\
         Top-1 Accuracy (MobileNetV2) & 65.2  & 68.6  & 70.7 & 71.2 & 71.5 \\ \hline
        Weight memory footprint & $1\times$ & $1\times$ & $1\times$ & $1.01\times$ & $0.96\times$ \\
         Top-1 Accuracy (ResNet18) & 70.0  & 67.8 &  70.7  & 70.9 &  71.0  \\
        \hline
        \end{tabular}
}
        \vspace{-5pt}
        \caption{   {Comparisons between PACT\citep{choi2018bridging}+UQ, PACT\citep{choi2018bridging}+PoT and DDQ on ImageNet. "DDQ (fixed)" and "DDQ (mixed)" indicate DDQ trained with fixed / mixed bitwidth. We see that DDQ+mixed surpasses all counterparts.
        }}
        \label{tab:mixed}
        \vspace{-10pt}
        
\end{table}

\begin{table*}
\centering
 
\scalebox{0.8}{
        \begin{tabular}{l |c | c c | c c }
        \hline
           & Full Precision & UQ & PoT & DDQ(fix) & DDQ(fix) + GradCorrect \\ \hline 
         Bitwidth (W/A )& 32 / 32 & 4 / 8 & 4 / 8 & 4 / 8 & 4 / 8 \\ \hline
         Top-1 Accuracy (MobileNetV2) & 71.9 & 67.1 & 69.2  & 71.2 & \textbf{71.6} \\
         Top-1 Accuracy (ResNet18) &  70.5 & 70.6 & 70.8 & 70.8 & \textbf{70.9} \\
         \hline\hline
        Bitwidth (W/A) &32 / 32 & 4 / 4 & 4 / 4 & 4 / 4 & 4 / 4 \\ \hline
         Top-1 Accuracy (MobileNetV2) & 71.9  &  65.2 & 68.4 & 70.4 & \textbf{70.7} \\
         Top-1 Accuracy (ResNet18) & 70.5 & 70.0  & 68.8  & 70.6  & \textbf{70.8} \\
        \hline\hline
        Bitwidth (W/A) & 32 / 32 & 2 / 4 & 2 / 4 & 2 / 4 & 2 / 4 \\ \hline
         Top-1 Accuracy (MobileNetV2) & 71.9 &  - & - &  60.1 & \textbf{63.7} \\
         Top-1 Accuracy (ResNet18) & 70.5 &  66.5  &  63.8 & 67.4  & \textbf{68.5} \\
        \hline\hline
        Bitwidth (W/A) & 32 / 32& 2 / 2 & 2 / 2 & 2 / 2 & 2 / 2 \\ \hline
         Top-1 Accuracy (MobileNetV2) & 71.9 & -  & -  & 51.1 & \textbf{55.4} \\
         Top-1 Accuracy (ResNet18) & 70.5 &  62.8  &  62.4 & 65.7  & \textbf{66.6} \\
        \hline 
        \end{tabular}
}   
        \vspace{-5pt}
        \caption{ {Ablation studies of  adaptive resolution and  gradient correction. ``UQ'' and ``PoT'' denote uniform and power-of-two quantization respectively. ``DDQ(fix)+GradCorrect'' refers to DDQ with gradient correction but fixed bitwidth. ``-'' denotes training diverged. ``4/8'' denotes training with 4-bit weights and 8-bit activations. Here we find that DDQ with gradient correction shows stable performances gains against DDQ (w/o gradient correction) and UQ/PoT baselines.}}
        \label{tab:level}
        \vspace{-15pt}
\end{table*}

\subsection{Ablation Study}

\textbf{Ablation Study I: mixed versus fixed precision.} 
Mixed-precision quantization is new in the literature and proven to be superior to their fixed bitwidth counterparts ~\citep{wang2019haq, uhlich2020mixed, cai2020rethinking, habi2020hmq}.  DDQ is naturally used to perform mix-precision training by a binary block-diagonal matrix $\U$.
In DDQ, each layer is quantized between 2-bit and a given maximum precision, which may be 4-bit / 6-bit / 8-bit.  Items of gate $\{\hat{g_i}\}_{i=1}^b$ are initialized all positively to $1e-8$, which means $\U$ is identity matrix and precision of layers are initialized to their maximum values. We set target bitwidth as $4$, constraining models to 4-bit memory footprint, and then jointly train $\{\hat{g_i}\}_{i=1}^b$ with other parameters of corresponding model and quantizers.
For memory constraints, $\alpha$ is set to $-0.02$ empirically. We use  learning rates $1e-8$ towards $\{\hat{g_i}\}_{i=1}^b$, ensuring sufficient training when precision decreasing. 
Fig.~\ref{fig:bit} depicts the evolution of bitwidth for each layer when quantizing a 4-bit ResNet18 using DDQ with maximum bitwidth $8$. As demonstrated, DDQ could learn to assign bitwidth to each layer, in a data-driven manner.

Table~\ref{tab:mixed} compares DDQ trained using mixed precision to different fixed-precision quantization setups, including DDQ with fixed precision, uniform (UQ) and power-of-two (PoT) quantization by PACT \citep{choi2018bridging} with gradient calibration \citep{jain2019trained, esser2019learned, jin2019towards, bhalgat2020lsq+}.

When the target bitwidth is 4, we see that DDQ trained with mixed precision significantly reduces accuracy drop of MobileNetv2 from 6.7\% (\eg PACT+UQ) to 0.4\%. See Appendix C.2 for more details.

\textbf{Ablation Study II: adaptive resolution.}
We evaluate the proposed adaptive resolution  by training DDQ with homogeneous bitwidth (\ie fixed $\U$) and only updating $\bm{q}$. 
%
%
%
%
%
Table~\ref{tab:level} shows performance of DNNs quantized with various quantization levels. We  see that  UQ and PoT incur a higher loss than DDQ, especially for MobileNetV2. We ascribe this drop to the irregular weight distribution as shown in Fig~\ref{fig:1}.  
Specially, when applying 2-bit quantization, DDQ still recovers certain accuracy compared to the full-precision model, while UQ and PoT may not converge. To our knowledge, DDQ is the first method to successfully quantize 2-bit MobileNet without using full precision in the activations.

\begin{figure}[t!]
\centering
\subfigure[Training dynamics of quantization error.]{
\vspace{-10pt}
\includegraphics[width=8cm]{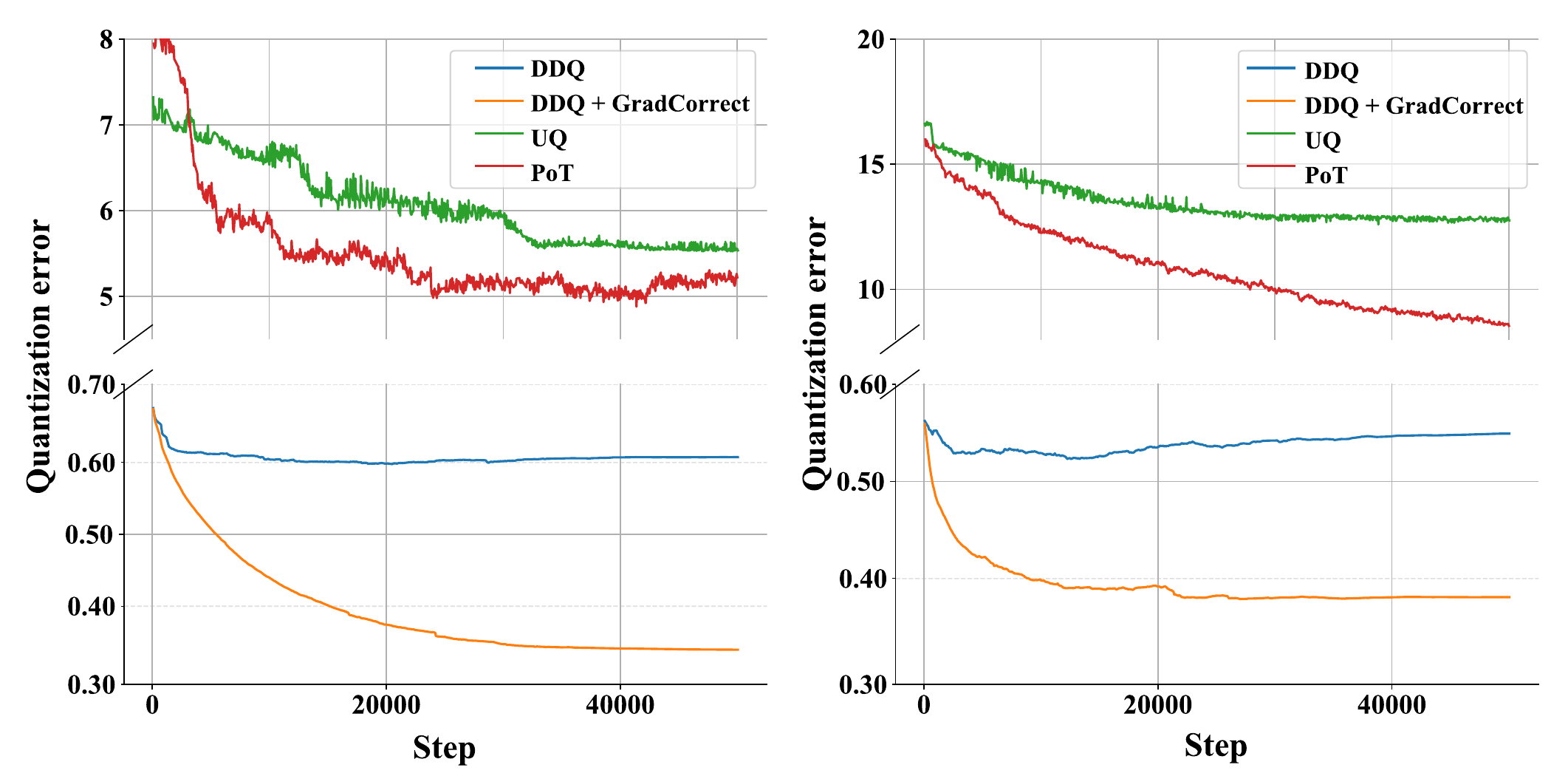}
\vspace{-10pt}
}
\centering
\subfigure[Learned quantization level for channels.]{
\vspace{-5pt}
\includegraphics[width=8cm]{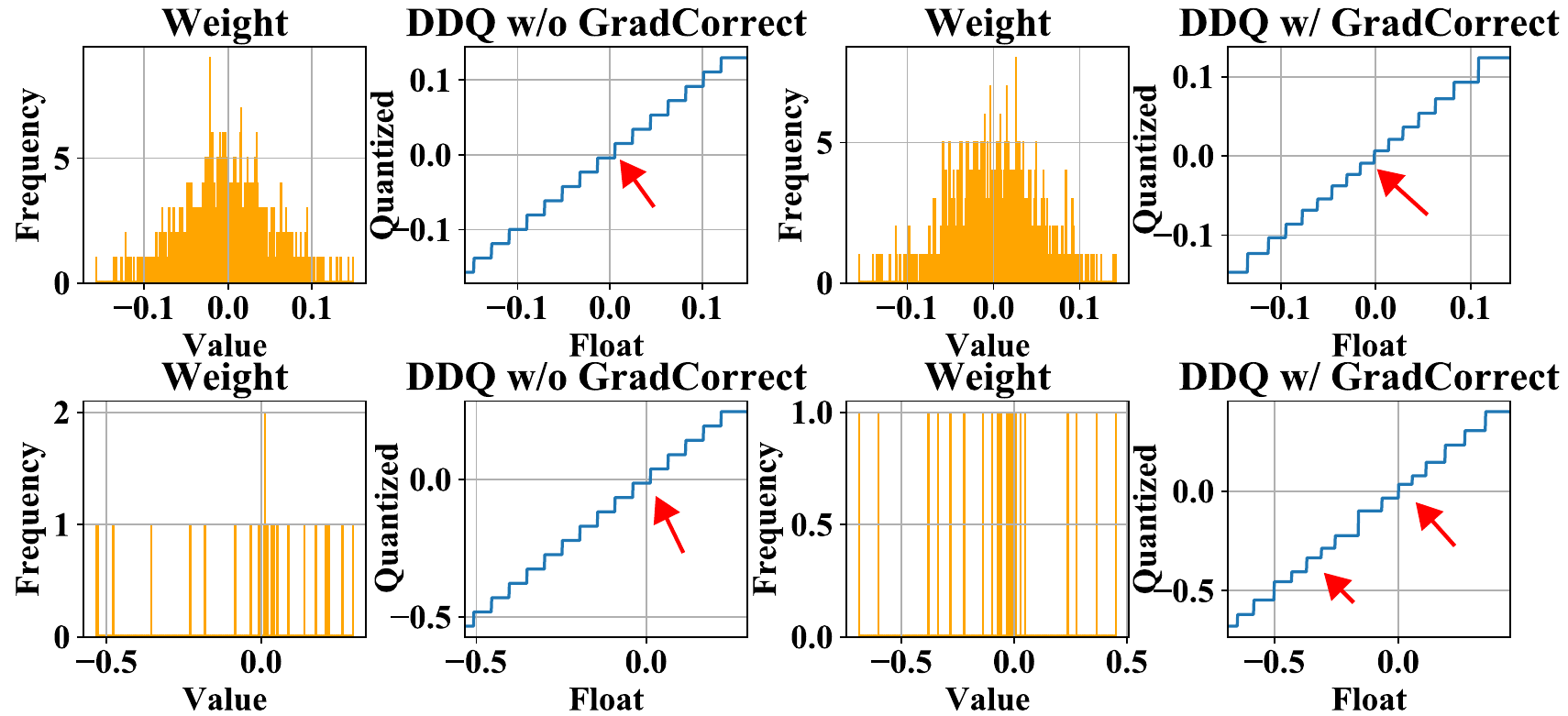}
\vspace{-10pt}
}
\vspace{-10pt}
 \caption{ {Training dynamics of quantization error 
 in MobileNetV2. (a) compares the quantization errors of PACT+UQ, PACT+PoT, and DDQ with/without gradient correction. DDQ with gradient correction shows stable convergence and lower quantization errors than counterparts.  (b) compares the converged quantization levels of DDQ for each channel with/without gradient correction, and dense regions are marked by arrows. Here we can also see that quantization levels learned by DDQ with gradient correction could fit original data distribution better. }}
 \label{fig:gradient}
 \vspace{-10pt}

\end{figure}

\textbf{Ablation Study III: gradient correction.}
We demonstrate how gradient correction improves DDQ.  
Fig~\ref{fig:gradient}(a) plots the training dynamics of layerwise quantization error \ie $ \| W_q - W \|_2^2$. 
%
We see that ``DDQ+gradient correction'' achieves low quantization error at each layer (average error is 0.35), indicating that the quantized values well approximate their continuous values.
Fig.~\ref{fig:gradient}(b) visualizes the trained quantization levels. DDQ trained with gradient correction would capture the distribution to the original weights, thus reducing the quantization error. As shown in Table.\ref{tab:level}, training DDQ  with gradient correction could apparently reduce accuracy drop from quantization.


\subsection{Evaluation on Mobile Devices}\label{app:mobile}
 In  Table \ref{tab:dsp}. we further evaluate DDQ on mobile platform to trade off of accuracy and latency.  
 With implementation stated in Appendix C.1, DDQ achieves over 3\% accuracy gains compared to UQ baseline. Moreover, in contrast to FP model, DDQ runs with about 40\% less latency with just a small accuracy drop (<0.3\%). 
 Note that all tests here are running under INT8 simulation due to the support of the platform. We believe the acceleration ratio can be larger in the future when deploying DDQ on more compatible hardwares.
\begin{table}
\centering
 
\scalebox{0.65}{
        \begin{tabular}{l| c c| c c }
        \toprule
        Methods & bitwidth(w/a) & Mixed-precision & Latency(ms) & Top-1(\%) \\ \hline
        FP & 32/32 &  & 7.8 & 71.9 \\
        UQ & 4/8 &  & 3.9 & 67.1 \\
        DDQ (fixed)$^1$ & 4/8 &  & 4.5 & 71.7 \\
        DDQ (mixed UQ)$^2$ & 4/8 & \checkmark & 4.1 & 70.8\\
        DDQ$^3$ & 4/8 & \checkmark & 5.1 & 71.9 \\
        \hline
        \end{tabular}

}
        \caption{  Comparison of Quantized MobileNetv2 runing on mobile DSPs. $^1$ Fixed precision DDQ. $^2$ Mixed precision DDQ with uniform quantizer constraints. $^3$ Original DDQ. ``w/a'' means the bitwidth for network weights and activations respectively.}
        \label{tab:dsp}
\end{table}

\section{Conclusion}
This paper introduced a differentiable dynamic quantization (DDQ), a versatile and powerful algorithm for training low-bit neural network, by automatically learning arbitrary quantization policies such as quantization levels and bitwidth. DDQ represents a wide range of quantizers. DDQ did well in the challenging MobileNet by significantly reducing quantization errors compared to prior work. 
%
%
We also show DDQ can learn to assign bitwidth for each layer of a DNN under desired memory constraints.
Unlike recent methods that may use reinforcement learning\citep{wang2019haq, yazdanbakhsh2018releq}, DDQ doesn't require multiple epochs of retraining, but still yield better performance compared to existing approaches. 
\vspace{-10pt}

\section{Acknowledgement}

This work is supported in part by Centre for Perceptual and Interactive Intelligence Limited, in part by the General Research Fund through the Research Grants Council
of Hong Kong under Grants (Nos. 14202217, 14203118, 14208619), in part by Research Impact Fund Grant No. R5001-18.
This work is also partially supported by the General Research Fund of HK No.27208720.

\newpage
\bibliography{references}
\bibliographystyle{icml2021.bst}

\end{document}